\documentclass[lettersize,journal]{IEEEtran}
\usepackage{amsmath,amsfonts}
\usepackage{amssymb}

\usepackage{array}
\usepackage[caption=false,font=normalsize,labelfont=sf,textfont=sf]{subfig}
\usepackage{textcomp}
\usepackage{stfloats}
\usepackage{url}
\usepackage{verbatim}
\usepackage{graphicx}
\usepackage{cite}
\usepackage{booktabs}

\usepackage{algorithm2e}
\usepackage{adjustbox}
\usepackage{multicol}
\usepackage{multirow}
\usepackage{pifont}
\newcommand{\cmark}{\ding{51}}
\newcommand{\xmark}{\ding{55}}
\usepackage[dvipsnames]{xcolor}
\graphicspath{ {./fig/} }
\begin{document}

\title{Single Stage Multi-Pose Virtual Try-On}

\author{Sen He, Yi-Zhe Song, \IEEEmembership{Senior Member,~IEEE}, Tao Xiang
\thanks{Sen He (e-mail: senhe752@gmail.com) is with Meta AI, London. This work was done when he was a postdoc at Centre for Vision Speech and Signal Processing (CVSSP), University of Surrey}
\thanks{Yi-Zhe Song and Tao Xiang are with the Centre for Vision Speech and Signal Processing (CVSSP), University of Surrey, Guildford GU2 7XH, United Kingdom.}

}

\markboth{Journal of \LaTeX\ Class Files,~Vol.~14, No.~8, September~2022}%
{Shell \MakeLowercase{\textit{et al.}}: A Sample Article Using IEEEtran.cls for IEEE Journals}


\maketitle

\begin{abstract}
 Multi-pose virtual try-on (MPVTON) aims to fit a target garment onto a person at a target pose. Compared to traditional virtual try-on (VTON) that fits the garment but keeps the pose unchanged, MPVTON provides a better try-on experience, but is also more challenging due to the dual garment and pose editing objectives. Existing MPVTON methods adopt a pipeline comprising three disjoint modules including a target semantic layout prediction module, a coarse try-on image generator and a refinement try-on image generator. These models  are trained separately, leading to sub-optimal model training and unsatisfactory results. In this paper, we propose a novel single stage model for MPVTON. Key to our model is a parallel flow estimation module that predicts the flow fields for both person and garment images conditioned on the target pose. The predicted flows are subsequently used to warp the appearance feature maps of the person and the garment images to construct a style map. The map  is then used to modulate the target pose's feature map for target try-on image generation. With the parallel flow estimation design, our model can be trained end-to-end in a single stage and is more computationally efficient, resulting in new SOTA performance on existing MPVTON benchmarks. We further introduce multi-task training and demonstrate that our model can also be  applied for traditional VTON and pose transfer tasks and achieve comparable performance to SOTA specialized models on both tasks.
\end{abstract}


\section{Introduction}

\begin{figure*}[t!]
    \centering
    \includegraphics[width=1.0\textwidth]{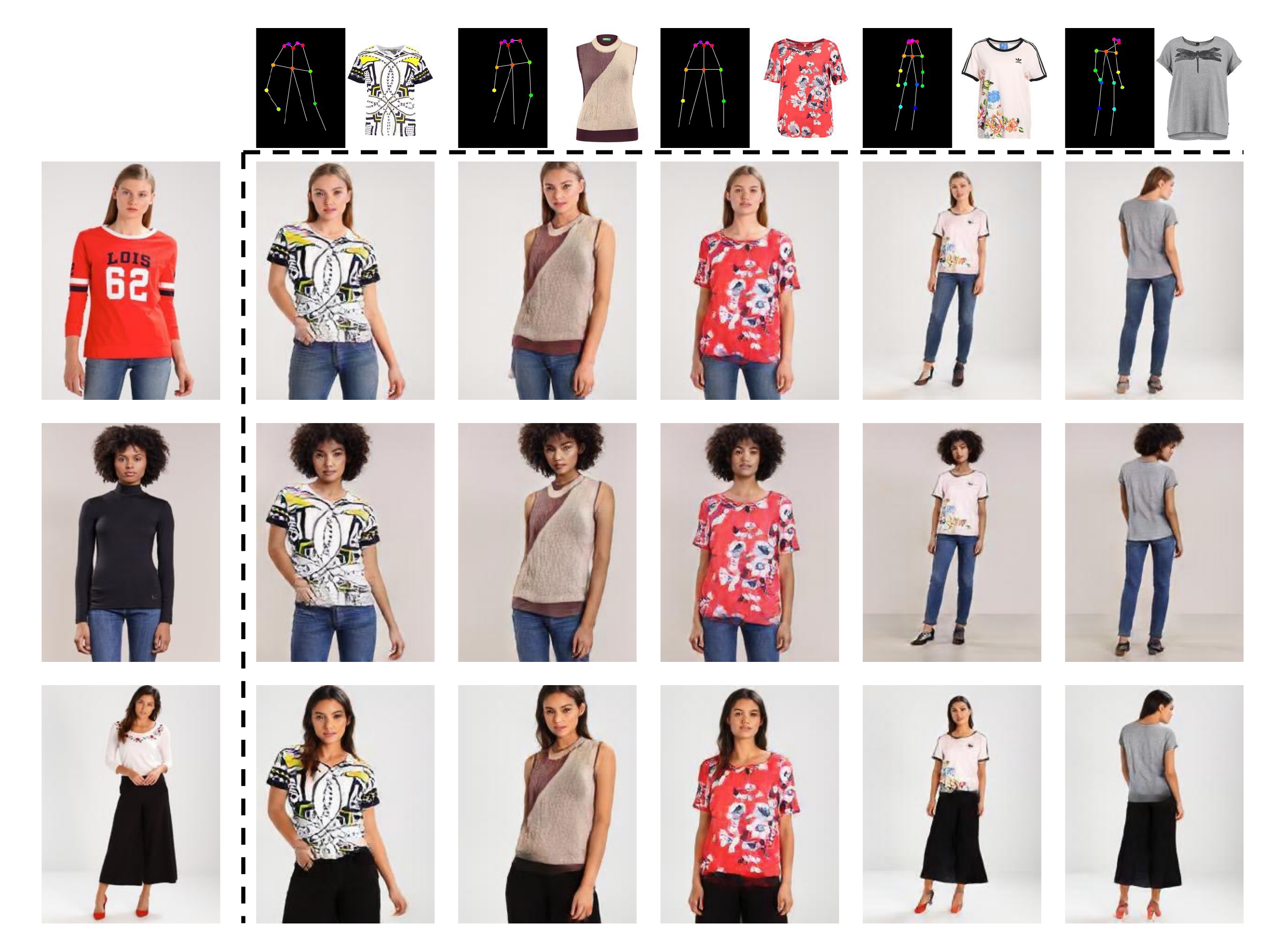}
    \caption{Given a person image, target pose and garment, our multi-pose virtual try-on model fits the garment onto the person at the target pose.}
    \label{fig:fig_pre}
\end{figure*}

Online apparel sales are soaring. The total U.S. retail apparel sales attributed to online shopping has grown from $26.6\%$ in 2018 to $46.0\%$ in 2020, and are expected to grow further. Though online apparel shopping provides a more convenient way to the shoppers, particularly during the pandemic, it also significantly increases the return cost of the retailers. Indeed, at least $30\%$ of all products ordered online are being returned as compared to $8.89\%$ in brick-and-mortar stores.  One of the key reasons for the return is the lack of a physic changing room to try-on the garment before purchase. To tackle this limitation, virtual try-on using computer vision has gained increasing attention recently \cite{han2018viton,wang2018toward,yu2019vtnfp,yang2020towards,issenhuth2020not,ge2021disentangled,han2019clothflow,ge2021parser,wang2020down,minar2020cp,lewis2021tryongan,xie2021towards}.

Traditional image-based virtual try-on fits a target garment into a person image at a fixed pose. This is usually done in two steps, i.e., garment warping and texture fusion. The first step warps the target garment to adapt to the human body in the person image. Then, in the second step, the texture from the warped garment and the non-garment region in the person image are fused  to generate the try-on image. Thanks to deep learning based techniques \cite{he2016deep,gong2017look,cao2017realtime,guler2018densepose,hinton2015distilling}, significant progresses have been made in this task in the past 5 years. However, traditional virtual try-on framework can only generate a try-on image at a fixed pose, which limits the virtual try-on experiences of an online shopper. This is because as how one often strikes different poses in front of a mirror when considering a garment item in a physical changing-room, virtually trying-on at different poses also helps the consideration. 

Multi-pose virtual try-on (MPVTON) \cite{dong2019towards,wang2020down} can fit a target garment into a person image and also transfer the pose of the person in the try-on image. It thus overcomes the limitation of the conventional try-on methods discussed above. However, MPVTON is also a much harder problem. This is because  it needs to simultaneously solve two challenging tasks, i.e., virtual try-on and pose transfer \cite{ma2017pose,dong2018soft,li2019dense,ren2020deep,zhang2021pise}. Each poses a different set of challenges and both are far from being solved.  As a result, MPVTON is much less studied.

Existing multi-pose virtual try-on methods \cite{dong2019towards,wang2020down} adopt a three-stage pipeline, in which target semantic layout, coarse and refined try-on images are generated sequentially by three different modules. Importantly the three modules are trained separately in the three stages. The reason for such a design is because previous methods need to firstly estimate the target semantic layout, which is then  used to either estimate the warping parameter of the target garment/source person \cite{dong2019towards,wang2020down} or guide the coarse try-on image generation \cite{wang2020down}. Due to the discrete nature of the target layout (i.e., semantic map, one-hot representation), it cannot be trained together with other modules by existing methods. The inability to train the three modules jointly results in sub-optimal model training and unsatisfactory try-on results. This also means that existing methods need a refinement module in the final stage to enhance their outputs, which increases the model complexity.

In this work, we propose a novel single stage framework for multi-pose virtual try-on. Instead of  predicting a target semantic layout to infer the warping parameters \cite{dong2019towards} or guide the coarse try-on image generation \cite{wang2020down},  the target pose is used as the condition to predict the flow fields to warp the target garment and the source person images. Our model adopts a   StyleGAN-like architecture \cite{karras2019style,karras2020analyzing}, which has shown to be effective on geometric transformation. For instance,  a single style vector has been  used to generate different viewpoints \cite{shen2021closed} of a face or deform the shape of a face \cite{or2020lifespan,he2021disentangled}. More specifically, in our model, conditioned on the target pose, a set of style vectors are  estimated  and then used to predict the flow fields for warping the feature maps extracted from both the target garment and source person images.  The warped appearance feature maps of garment and person images are used for style modulation in a SPADE  generator \cite{park2019semantic}  to generate the final target try-on image. Our model is single stage and can be trained end-to-end. Even without a refinement module, our model can achieve state-of-the-art performance on multi-pose virtual try-on benchmark \cite{dong2019towards}. This single stage design also brings  flexibility and versatility: we further design a multi-task training objective for our  model to perform three tasks, i.e., multi-pose virtual try-on, pose transfer and virtual try-on. This enables it to also perform on par with SOTA methods specifically designed for one of the other two auxiliary tasks.

\textbf{The contributions} of this work are as follows: (1) We propose a novel single stage multi-pose virtual try-on model. Our model can be trained end-to-end and is more computationally efficient. (2) We design a multi-task learning objective to train our model. This learning objective further enhances our model's performance and enables it to perform well on traditional virtual try-on and pose transfer tasks. (3) Extensive experiments are carried out to demonstrate the effectiveness of our proposed methods: Our model achieves new state-of-the-art performance on existing multi-pose virtual try-on benchmarks. 

\section{Related Work}

\paragraph{\textbf{Pose transfer}} Pose transfer aims to transfer the pose of a person in the image to another pose while preserve the person's appearance. The pioneer work \cite{ma2017pose} proposes a two stages framework to  first generate a coarse image at target pose and then refine the appearance in the second stage. More advanced methods directly predict the deformation fields between the source pose and the target pose, and then use it to warp the person in the source image. \cite{siarohin2018deformable} proposes to split the human body into several parts and then predicts the affine transformation between each part in the source person's pose and target pose. \cite{dong2018soft} proposes to predict the target semantic layout and then use a matching network to estimate the deformation field between the source and target semantic layout.
\cite{ren2020deep} directly uses a fully convolutional neural network to predict the deformation field. Recently, \cite{zhang2021pise,ma2021must} apply the idea of feature modulation \cite{karras2019style,karras2020analyzing} for pose transfer, where they extract the style of each semantic region from the source image and then use it to modulate the feature in the target semantic layout.

Most pose transfer models \cite{ma2017pose,siarohin2018deformable,dong2018soft,ren2020deep} are not designed also tackle the multi-pose virtual try-on task. A couple of recent models \cite{zhang2021pise,ma2021must} could be extended for multi-pose virtual try-on, e.g., they can directly extract the style vector from the target garment image and then use it to modulate the features from the garment region in the target image. However, these methods cannot preserve the texture details of the target garment, as they embed all details in a single vector \cite{kim2021exploiting}.

\paragraph{\textbf{Image based virtual try-on}}  Existing methods for image based virtual try-on can be divided into two groups, namely parser based and parser free methods.
Parser based model first de-cloth the person image (mask out the garment region in the person image) with an off-the-shelf human parser, e.g., \cite{gong2017look}. They then use the de-clothed person image to estimate a garment warping parameter \cite{han2018viton} or flow field \cite{han2019clothflow} to warp the target garment. Later on the warped garment and the de-clothed person are fused together to generate the final try-on image. On the other hand, 
parser free methods \cite{issenhuth2020not,ge2021parser} first train a parser based model and then distill a parser free model from the pre-trained parser based model. One advantage of parser free models is that they do not need to use a human parser during inference, therefore avoiding being negatively impacted by parsing errors.

However, prior image based virtual try-on methods can only generate a single try-on image at the person's original pose, while in a real-world application, multi-pose try-on is more desirable to shoppers. The only exception as far as we know is \cite{cui2021dressing} which sequentially generates the try-on images at multiple poses. Our model differs from  \cite{cui2021dressing} in two aspects: (1) Our model during inference only needs a single forward pass to generate the multi-pose try-on image. (2) We use a novel style-based flow field estimation for warping both the person image and the garment image. In contrast, a U-Net architecture is adopted in \cite{cui2021dressing}.  Our experiments show that that our style-based model achieves better performance.

\paragraph{\textbf{3D virtual try-on}} 3D virtual try-on methods \cite{bhatnagar2019multi,mir2020learning} represent  human body with a 3D parametric model \cite{loper2015smpl} and then mesh the target garment to render the corresponding region on the human body. As pose control can be easily achieved with a parametric body model, a  3D virtual try-on model in theory can also be applied to generate multi-pose try-on images: after try-on, one can first manipulate the 3D parametric human body for pose transfer and then project it to the 2D image plane. However, collecting large scale 3D datasets is expensive  and  laborious,  thus  posing  a  constraint  on  the scalability of a 3D virtual try-on model. Moreover, this paper only focus on key points based pose representation \cite{cao2017realtime}, which is not directly comparable with a 3D human body model that requires a much detailed pose representation.
\paragraph{\textbf{StyleGAN}} StyleGAN architecture \cite{karras2019style,park2019semantic,karras2020analyzing} has achieved great successes in both image generation and manipulation \cite{he2019attgan,shen2020interpreting}. Its successes rely on its smart noise-to-style architecture (eight fully connected layers) and the learned highly disentangled latent style space. In this work, StyleGAN architecture is used for both flow filed estimation and image generation.

\section{Methodology}

\subsection{Problem definition}

\begin{figure*}[t!]
    \centering
    \includegraphics[width=1.0\textwidth]{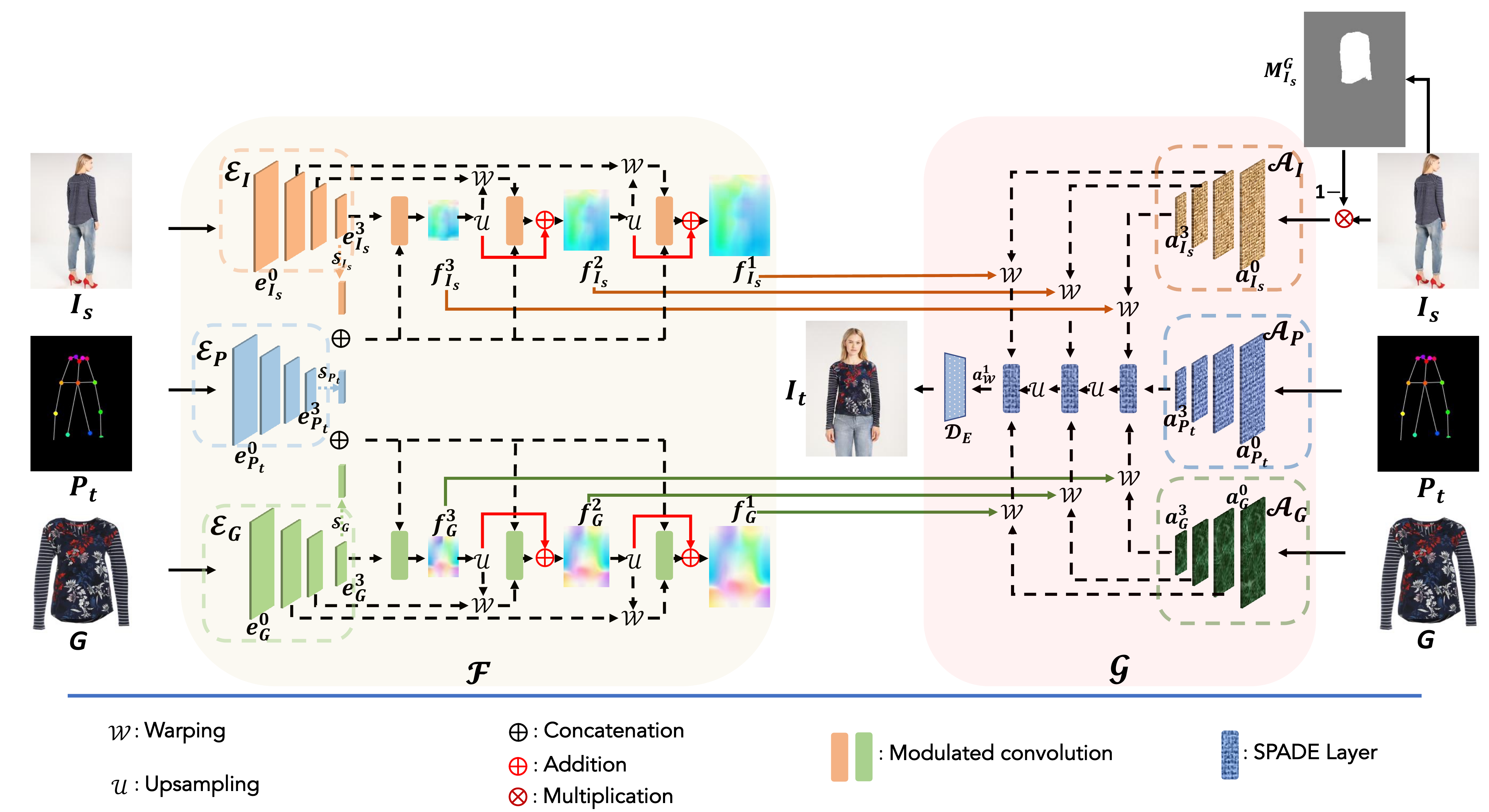}
    \caption{A schematic of our framework. The flow estimation module $\mathcal{F}$ estimates the flow fields for the input person and garment images. The image generation module $\mathcal{G}$ uses the predicted flow fields to warp the appearance feature maps of garment and masked person images for target try-on image generation.}
    \label{fig:fig_archi}
\end{figure*}

Given a person image ($I_{s} \in \mathbb{R} ^ {3 \times H \times W}$), a target garment image ($G \in \mathbb{R} ^ {3 \times H \times W}$), and a target pose ($P_{t} \in \mathbb{R} ^ {18 \times H \times W}$, denotes 18 human body joints \cite{cao2017realtime}), the goal of multi-pose virtual try-on is to generate a try-on image ($I_{t} \in \mathbb{R} ^ {3 \times H \times W}$) where the garment in $G$ fits to the person in $I_{s}$ at the target pose $P_{t}$.

Our model, as illustrated in Fig.~\ref{fig:fig_archi}, consists of two modules, i.e., flow estimation module ($\mathcal{F}$) and image generation module ($\mathcal{G}$). The flow estimation module takes the target pose, person image and garment image as inputs, and generate two flow fields for warping the person and garment respectively. The image generation module takes as inputs a masked person image, the target pose, the garment image and the estimated flow fields, and generate the target try-on image. The whole process in our model can be defined as:
\begin{equation}
    I_{t} = \mathcal{G}(P_{t}, I_{s} \cdot (1-M_{I_{s}}^{G}), G, \mathcal{F}(P_{t}, I_{s}, G)).
\end{equation}
Each module will be detailed in the following sections.

\subsection{Flow fields estimation}

As neither the person nor the target garment is spatially aligned with the target pose, our first step is to use $\mathcal{F}$ to predict the warping flow fields for the person and garment  respectively (${\{f_{Is}^{i}\}_{1}^{N}}$ and $\{f_{G}^{i}\}_{1}^{N} \in \mathbb{R}^{2 \times h_{i} \times w_{i}}$, $N=3$ in Fig.~\ref{fig:fig_archi} for simplicity). With the predicted flow fields, we can then warp the feature maps extracted from  the person and garment images, so that they are spatially aligned with the target pose.

Instead of relying the predicted target layout  to estimate the flow fields as in existing methods \cite{dong2019towards,wang2020down}, we directly use the target pose as a conditioning style to modulate the feature maps of the person and garment images, in order to  generate the flow fields. Our method is motivated by the existing style-based image geometric editing models, where a single style vector can control different viewpoints or shapes of a face \cite{he2021disentangled,or2020lifespan,shen2021closed}. Here we directly use a style vector to predict the deformative flow fields.

Concretely, we use three encoders to extract the structure features of person image, garment image and target pose:

\begin{equation}
    \begin{aligned}
         &\{e_{Is}^{i}\}_{0}^{N}=\mathcal{E}_{I}(I_{s}),\\
        &\{e_{G}^{i}\}_{0}^{N}=\mathcal{E}_{G}(G), \\ &\{e_{Pt}^{i}\}_{0}^{N}=\mathcal{E}_{P}(P_{t}),
    \end{aligned}
\end{equation}
where $e_{Is}^{i}, e_{G}^{i} \ \text{and} \ e_{Pt}^{i} \in \mathbb{R}^{c_{i} \times h_{i} \times w_{i}}$ are the structure feature maps of person, garment and target pose respectively. Then, we construct two flow estimation style vectors ($s_{Is} \ \text{and} \ s_{G} \in \mathbb{R}^{2c_{N}}$) for person and garment from the lowest resolution feature maps:
\begin{equation}
    \begin{aligned}
        &s_{Is}=[\mathcal{S}_{Pt}(e_{Pt}^{N}), \mathcal{S}_{Is}(e_{Is}^{N})], \\
        &s_{G} = [\mathcal{S}_{Pt}(e_{Pt}^{N}), \mathcal{S}_{G}(e_{G}^{N})],
    \end{aligned}
\end{equation}
where $\mathcal{S}_{Pt}, \mathcal{S}_{Is}, \text{and} \ \mathcal{S}_{G}$ are style embedding blocks\footnote{These blocks project the lowest structure feature maps into vectors.}.

With the constructed style vectors, we can predict the flow fields at different resolutions via modulated convolution \cite{karras2020analyzing}. Specifically, for the structure feature map at the $i^{th}$ resolution, it is firstly warped by the predicted flow field at the preceding resolution, and then used to predict a residual flow:
\begin{equation}
    \begin{aligned}
        &f_{Is}^{i} = conv_m(\mathcal{W}(e_{Pt}^{i},\ \mathcal{U}(f_{Is}^{i+1})),s_{Is}) + \mathcal{U}(f_{Is}^{i+1}), \\ 
        &f_{G}^{i} = conv_m(\mathcal{W}(e_{G}^{i},\ \mathcal{U}(f_{G}^{i+1})),s_G) + \mathcal{U}(f_{G}^{i+1}),
    \end{aligned}
\end{equation}
where $conv_{m}$ is a modulated convolution, $\mathcal{W} \ \text{and} \ \mathcal{U}$ denote the warping and upsampling operations respectively. For the lowest resolution structure feature maps ($e_{Is}^{N} \ \text{and} \ e_{G}^{N}$), they are directly used to predict the lowest resolution flows without warping. Note that we do not predict flows at the resolution of input image as we need to keep some degree of abstractness of the feature maps for image generation when hallucination is needed for large pose changes\footnote{Full resolution appearance feature maps usually capture low-level features, thus are not suited for appearance hallucination, when e.g., the input person  is of a frontal view but the target pose is of a back view.}.

\subsection{Image generation}

The image generation module $\mathcal{G}$ receives the estimated flow fields from $\mathcal{F}$ and uses it to warp the appearance feature maps of garment image and the non-garment regions of the person image for target try-on image generation.

Concretely, we use another three encoders to extract the appearance features of the non-garment regions in person image ($\{a_{Is}^{i}\}_{0}^{N}$), target garment image ($\{a_{G}^{i}\}_{0}^{N}$) and the latent embedding of the target pose ($\{a_{Pt}^{i}\}_{0}^{N}$) respectively:
\begin{equation}
\begin{aligned}
     \{a_{Is}^{i}\}_{0}^{N}&=\mathcal{A}_{I}(I_{s}\cdot (1-M_{I_{s}}^{G})),\\
    \{a_{G}^{i}\}_{0}^{N}&=\mathcal{A}_{G}(G),\\ 
    \{a_{Pt}^{i}\}_{0}^{N}&=\mathcal{A}_{P}(P_{t}),
\end{aligned}
\end{equation}
where $M_{I_{s}}^{G}$ is the mask for the garment region in the person image, extracted by an off-the-shelf human parser \cite{gong2017look}. Note that the three encoders in $\mathcal{G}$ are not shared with those in $\mathcal{F}$, as flow estimation and image generation requires quite different features\footnote{Flow estimation usually depends on structure features to find the correspondence, while image generation needs texture features to reconstruct the appearance of person and garment.}.

Then, we use $a_{Pt}^{N}$ as the initial embedding for the target try-on image and gradually modulate it via SPADE \cite{park2019semantic} layers with the warped person and garment features. Specifically, at each resolution, the SPADE layer takes as inputs the modulated feature map ($a_{\mathcal{W}}^{i+1}$) in the preceding layer and the concatenation of warped person and garment appearance features, and output a new modulated feature map:
\begin{equation}
    a_{\mathcal{W}}^{i} = \text{SPADE}(\mathcal{U}(a_{\mathcal{W}}^{i+1}), [\mathcal{W}(a_{Is}^{i}, f_{Is}^{i}),\mathcal{W}(a_{G}^{i}, f_{G}^{i})]).
\end{equation}
Within the SPADE layer, the upsampled input $a_{\mathcal{W}}^{i+1}$ is first normalized by its own mean and variance and then location-wisely transformed by the affine transformation parameters generated by the concatenated feature maps.

The last modulated feature map $a_{\mathcal{W}}^{1}$ is fed into a decoding block\footnote{$\mathcal{D}_{E}$ consists of convolution, upsampling and activation layers.} ($\mathcal{D}_{E}$) for the final try-on image generation:
\begin{equation}
    I_{t} = \mathcal{D}_{E}(a_{\mathcal{W}}^{1}).
\end{equation}
\subsection{Multi-task learning}

We further design a multi-task pipeline to train our model, i.e., we train our model to simultaneously perform three tasks: the main multi-pose virtual try-on  task and two auxiliary tasks, namely traditional virtual try-on and pose transfer. Due to simple one-stage design,  our model is flexible in training for different tasks. Concretely, for the traditional virtual try-on task, we replace the target pose with the original pose in the person image. For the pose transfer task, we replace the garment image with the garment segmented from the person image. The traditional virtual try-on requires the model to predict a zero flow fields for the person image. Meanwhile,  pose transfer needs the model to extract more robust feature from the warped garment on the person. Having them as auxiliary tasks thus help our model learn features which will benefit both $\mathcal{F}$ and $\mathcal{G}$.

\subsection{Learning objectives}
Ideally, a MPVTON model should be trained with paired data containing the same person wearing different garments at different poses. Instead, the existing multi-pose virtual try-on benchmark \cite{dong2019towards} only has the same person wearing the same garment at different poses. Therefore our model is trained with a reconstruction objective. During training, the input garment is exactly the same as the one in the person image, and we apply a perceptual loss between the generated try-on image $I_{t}$ and the ground truth image $I_{gt}$:
\begin{equation}
    L_{p} = \sum_{i} \lVert \phi_{i}(I_{t}) - \phi_{i}(I_{gt}) \rVert,
\end{equation}
where $\phi_{i}$ is the $i^{th}$ block of the pre-trained VGG network \cite{simonyan2014very}.

To supervise the training of the flow estimation module $\mathcal{F}$, we apply correction losses ($L_{Ic} \ \text{and} \ L_{Gc}$) \cite{ren2020deep} to the estimated flow fields such that the warped person and garment images are highly correlated with the ground truth person image and garment regions in the ground truth image, respectively:
\begin{equation}
\begin{aligned}
    L_{Ic} &= \sum_{i}L_{Ic}^{i},L_{Gc} = \sum_{i}L_{Gc}^{i},
\end{aligned}
\end{equation}
and,
\begin{equation}
    \begin{aligned}
    L_{Ic}^{i} &= \frac{1}{N}\sum_{j}exp(-\frac{\mu(\phi_{l}^{j}(\mathcal{R}(I_{gt})),  \phi_{l}^{j}(\mathcal{W}(\mathcal{R}(I_{s}),f_{Is}^{i})))}{\mu_{max}^{I_{i}}}), \\
    L_{Gc}^{i} &= \frac{1}{N}\sum_{j}exp(-\frac{\mu(\phi_{l}^{j}(\mathcal{R}(I_{gt} \cdot M_{I_{gt}}^{G})), \phi_{l}^{j}(\mathcal{W}(\mathcal{R}(G),f_{G}^{i})))}{\mu_{max}^{G_{i}}}),
\end{aligned}
\end{equation}
where $\phi_{l}^{j}(\cdot)$ denotes the feature at the $j^{th}$ location from the $l^{th}$ layer of a pretrained VGG network\footnote{We follow \cite{ren2020deep} to select $l$ for $i^{th}$ flow.}, $\mathcal{R}(\cdot)$ is the resizing operation which resizes its input to match the $i^{th}$ flow resolution, $\mu(\cdot,\cdot)$ is the cosine similarity operator, $\mu_{max}^{I_{i}} \ \text{and} \ \mu_{max}^{G_{i}}$ are the maximum cosine similarity values between the feature maps of the warped image and the target image, and $M_{I_{gt}}^{G}$ is the mask of garment region in the ground truth image.

Furthermore,  a conditional adversarial loss is used to improve the realism of the generated images (for simplicity, the discriminator, $D$, is not illustrated in Fig.~\ref{fig:fig_archi}):
\begin{equation}
    L_{adv} =  \  \mathbb{E}_{I_{gt}}[log(D(I_{gt} | P_{t})] + \mathbb{E}_{I_{t}}[1-log(D(I_{t} | P_{t})].
\end{equation}

The overall training objectives are:
\begin{equation}
    L = \lambda_{p}L_{p} + \lambda_{c}(L_{Ic} + L_{Gc}) + \lambda_{adv}L_{adv},
\end{equation}
where $\lambda_{p}, \lambda_{c} \ \text{and} \ \lambda_{adv}$ are hyperparameters for balancing the three objectives.

\subsection{Inference}

During inference, for each identity, we randomly sample a garment image and a target pose from other identities in the dataset. When our model is applied to traditional virtual try-on task, the target pose is replaced by the original pose in the input person image. And for pose transfer, the target garment is replaced by the segmented garment in the input person image. 

\section{Experiments}

\begin{figure*}[t!]
    \centering
    \includegraphics[width=1.0\textwidth]{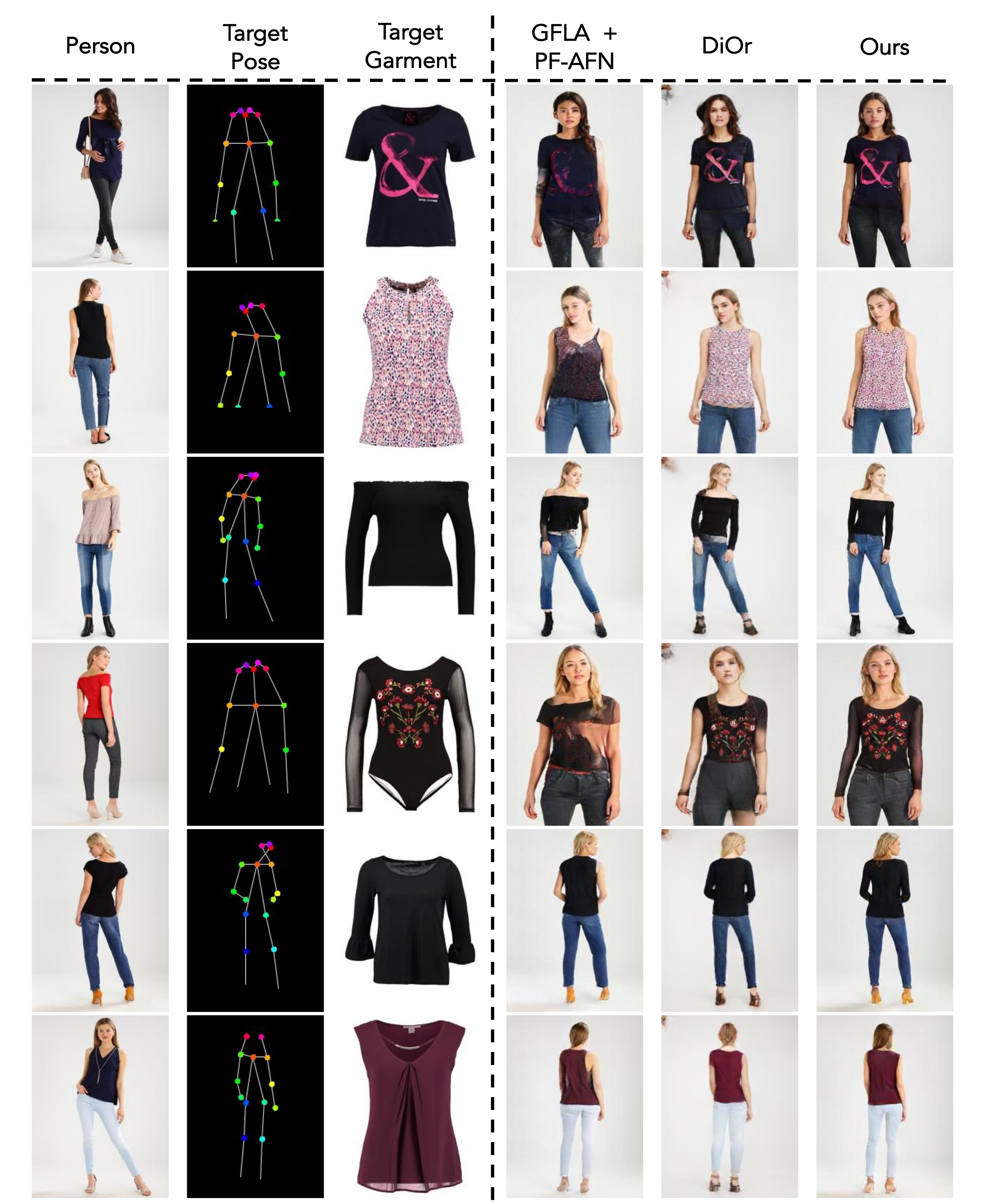}
    \caption{Qualitative results from different models (GFLA \cite{ren2020deep} + PF-AFN \cite{ge2021parser}, DiOr \cite{cui2021dressing} and ours) on the MPV test split.}
    \label{fig:fig_main_results}
    \vspace{-0.4cm}
\end{figure*}

\paragraph{\textbf{Dataset}} We experiment our model on the MPV \cite{dong2019towards} dataset. It is the only publicly available benchmark for multi-pose virtual try-on at present. MPV contains 35,687 person images and 13,524 garment images. Each person in MPV has at least two different poses wearing the same garment. As per \cite{dong2019towards}, we constructed 62,780 tuples where each contains two person images (the same person wearing the same garment at different poses) and one garment image. These tuples are further split into 52,236 for training and 10,544 for testing. The image resolution is set as $256 \times 192$.

\paragraph{\textbf{Implementation details}} Our model is implemented in PyTorch. We train our model with a single Nvidia RTX2080-Ti GPU. The batch size is set to 6. As per standard \cite{ren2020deep,cui2021dressing}, we first train the flow estimation module $\mathcal{F}$ with 200k iterations as a warm up process and then train the whole model with another 600k iterations. We train our model with the Adam optimizer \cite{kingma2014adam}. The initial learning rate is set to $5e-4$ which is linearly decayed after 300k iterations.

\paragraph{\textbf{Evaluation metrics and baselines}} We evaluate our model both automatically and manually (human study). In automatic evaluation, as per standard in multi-pose virtual try-on, we evaluate model performance using structure similarity (SSIM) \cite{wang2004image}, masked structure similarity (mask-SSIM, only evaluate the SSIM for the foreground person region), garment structure similarity (garment-SSIM, only evaluate the SSIM for the garment region in the try-on image),  learned perceptual image patch similarity (LPIPS) \cite{zhang2018unreasonable} and Fr\'{e}chet Inception Distance (FID) \cite{heusel2017gans}. According to \cite{rosca2017variational,ge2021parser}, inception score (IS) \cite{salimans2016improved} is not suitable for evaluating virtual try-on images; it is thus not used in this work. In the manual (subjective) evaluation, we run perceptual study on Amazon Mechanical Turk (AMT) to compare the quality of the generated multi-pose try-on images from different models. Given an input person image, a garment image, a target pose and the generated  try-on  image  from  two compared models,  the  AMT  workers were asked to vote on which generated try-on image is better. Each AMT worker was randomly allocated 100 images to compare two models.  10 workers participated in the evaluation for all models comparison.

We compare our method with previous multi-pose virtual try-on models CP-VTON \cite{wang2018toward} (re-purposed for multi-pose virtual try-on \cite{wang2020down}), MG-VTON \cite{dong2019towards} and DLD \cite{wang2020down}. We also compare with two hybrid models, PF-AFN \cite{ge2021parser} + GFLA \cite{ren2020deep} and GFLA + PF-AFN\footnote{For fair comparison, we trained PF-AFN without distillation on MPV. PF-AFN + GFLA means that we first do virtual try-on use PF-AFN and then do pose transfer with GFLA. GFLA + PF-AFN means the opposite order.}, which combine independently trained state-of-the-art virtual try-on (PF-AFN) and pose transfer (GFLA) models in different orders. Finally, we also compare with a recent virtual try-on model, DiOr \cite{cui2021dressing}, that can also be applied for multi-pose virtual try-on.

\paragraph{\textbf{Main results}} The quantitative results on MPV testing split are shown in Table \ref{tab:mp-vton}. It can be seen that our model achieves new state-of-the-art  performance. Our model outperforms previous models  across all five metrics with the only exception of DLD on mask-SSIM\footnote{DLD is evaluated on a subset of simple positive perspective images}. In the manual evaluation, our model also outperforms most compared methods by more than $10\%$ preference rate, which is consistent with the results in the automatic evaluation metrics. It is interesting to note that for the hybrid models (PF-AFN + GFLA and GFLA + PF-AFN), the try-on/pose transfer order plays an important role. Firstly transferring the persons pose with the pose transfer model (GFLA) followed by fitting the target garment with the virtual try-on model (PF-AFN) is significantly better than the other way around. Some qualitative results from different models are shown in Fig.~\ref{fig:fig_main_results}. Overall, our method can generate better try-on images in terms of realism of images, better garment fitting, preservation of garment texture detail and person identity, and less artifacts. Crucially, our model learned to hallucinate the back view of garments given only its front view (see examples in the last two rows of Fig.~\ref{fig:fig_main_results}).

We further compare the efficiency of different models. We run all models with the same hardware\footnote{A single NVIDIA-2080 Ti GPU.} and test their inference memory cost and time consumption. The results are shown in Table~\ref{tab:complexity}. It is clear that single stage models significantly outperform multi-stage models and our model has the lowest running time. Note that DiOr \cite{cui2021dressing} directly use the garment segmented from person image and does not need a separate garment encoder to extract the appearance feature of the garment image; it thus requires less inference memory.

\begin{table*}[t]
    \centering
    \caption{Quantitative results of different multi-pose virtual try-on models. \\ $\star$ means only a subset of positive perspective images in MPV was used in the training and testing. PR: preference rate in manual evaluation (other models/our model).}
    \begin{tabular}{c| c c c c c|c }
    \toprule
    Methods& SSIM $\uparrow$& mask-SSIM $\uparrow$ & garment-SSIM $\uparrow$ & FID $\downarrow$& LPIPS $\downarrow$ & PR \\
    \midrule
    CP-VTON \cite{wang2018toward}& 0.563&0.548&0.583 & 38.19&0.248&-\\
    MG-VTON \cite{dong2019towards}&0.705&0.776& 0.721& 22.42&0.202&-\\
    $\text{DLD}^{\star}$ \cite{wang2020down}&0.723&\textbf{0.784}&0.795 &16.01& 0.187&45.7\% / 54.3\% \\
    PF-AFN \cite{ge2021parser} + GFLA \cite{ren2020deep}&0.662&0.701&0.717 &17.04&0.263&37.6 \% / 62.4\%\\
    GFLA \cite{ren2020deep} + PF-AFN \cite{ge2021parser}&0.722&0.749&0.754 &16.03&0.218&41.7\% / 58.3\%\\
    DiOr \cite{cui2021dressing}&0.748&0.770& 0.782&13.15&0.188&45.2\% / 54.8\%\\
    \midrule
    Ours&\textbf{0.761}&0.782&\textbf{0.807} &\textbf{9.34}&\textbf{0.170}&-\\
    \bottomrule
    \end{tabular}
    
    \label{tab:mp-vton}
\end{table*}

\begin{table}[t]
    \centering
    \caption{Inference memory cost and time consumption (batch size = 1) of different multi-pose virtual try-on methods. \\ IM: Inference Memory, IT: Inference Time.}
    \setlength{\tabcolsep}{3pt}
    \begin{tabular}{c| c | c c }
    \toprule
    Methods& Single stage& IM & IT  \\
    \midrule
    
    {DLD} \cite{wang2020down}& \xmark &1625MB&0.10s\\
    GFLA \cite{ren2020deep} + PF-AFN \cite{ge2021parser}& \xmark &2818MB&0.13s\\
    DiOr \cite{cui2021dressing}& \cmark &\textbf{1175MB}&0.07s\\
    \midrule
    Ours& \cmark &1275MB&\textbf{0.03s}\\
    \bottomrule
    \end{tabular}
    
    \label{tab:complexity}
\end{table}

\begin{table}[t]
    \centering
    \caption{Ablation studies of our proposed method}
    \setlength{\tabcolsep}{5pt}
    \begin{tabular}{c c c | c c}
    \toprule
         single stage& shared encoders &multi-task training& SSIM $\uparrow$& FID $\downarrow$  \\
         \midrule
         \xmark&\cmark&\xmark&0.732&13.19\\
         \cmark&\cmark&\xmark&0.741&12.07\\
         \cmark&\cmark&\cmark&0.747&11.38\\
         \cmark&\xmark&\xmark&0.753&10.53\\
         \cmark&\xmark&\cmark&0.761&9.34\\
    \bottomrule
    \end{tabular}
    \label{tab:ablaation}
\end{table}

\paragraph{\textbf{Ablation study}} Here we validate the design of our proposed method. We mainly validate three designs. (1) Single stage training vs. multi-stage training. In single stage training, we jointly train our flow estimation module $\mathcal{F}$ and image generation module $\mathcal{G}$, while in multi-stage training, we first train the $\mathcal{F}$ and then fix it when training $\mathcal{G}$.  (2) Shared encoders vs. non-shared encoders. We investigate whether we need to share the encoders for person image, target garment and target pose between $\mathcal{F}$ and $\mathcal{G}$. (3) Multi-task training. We examine whether the multi-task training (multi-pose virtual try-on + traditional virtual try-on + pose transfer) improves our model performance.

The main results are shown in Table~\ref{tab:ablaation}. It can be seen that single stage training improves the model performance. This validate our assumption that by having a single stage design, we can avoid the sub-optimal training suffered by the existing multi-stage  models \cite{dong2019towards,wang2020down}. By using non-shared encoders between $\mathcal{F}$ and $\mathcal{G}$, both SSIM and FID score are improved. This indicates that flow estimation and image generation indeed need different features from non-shared encoders. Finally,  multi-task learning further boosts our model's performance.

\paragraph{\textbf{Traditional virtual try-on and pose transfer}} In this experiment, we evaluate our model on traditional virtual try-on task where the person's pose is fixed and pose transfer task where the garment is fixed. For traditional virtual try-on, we directly evaluate our model on the popular VITON benchmark \cite{han2018viton} and compare with recent SOTA models VTON \cite{han2018viton}, CP-VTON \cite{dong2019towards}, ACGPN \cite{yang2020towards} and PF-AFN \cite{ge2021parser}. For pose transfer. we evaluate our model on MPV and DeepFashion\footnote{When evaluating on DeepFashion, we directly use our model trained on MPV to generate test images without fine-tuning.} \cite{liu2016deepfashion} test splits and compare with SOTA pose transfer models ADGAN \cite{men2020controllable}, GFLA \cite{ren2020deep} and DiOr \cite{cui2021dressing}.

\begin{figure}[h!]
    \centering
    \includegraphics[width=0.5\textwidth]{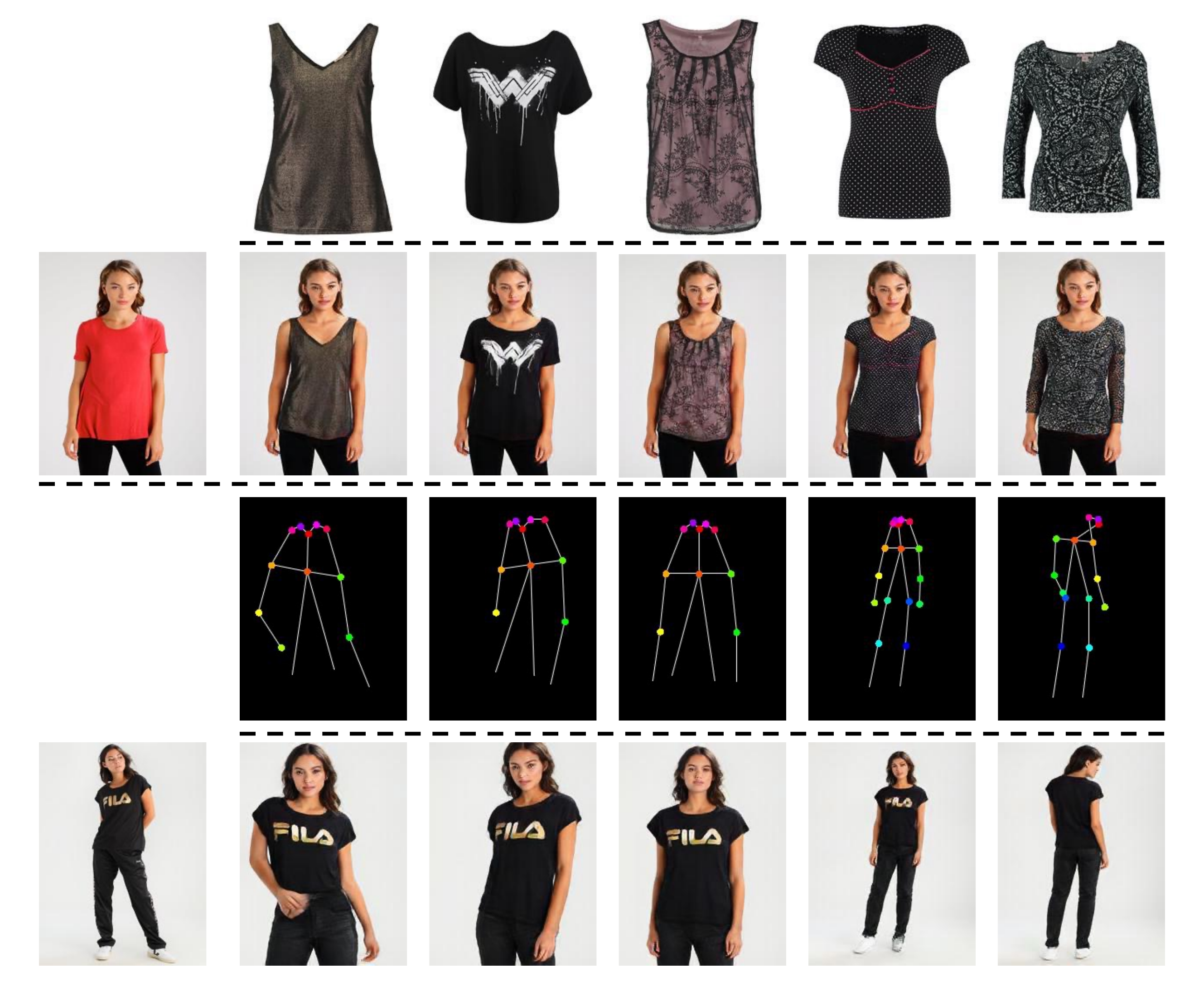}
    \caption{Our model's qualitative results on traditional virtual try-on (top two rows) and pose transfer (bottom two rows).}
    \label{fig:fig_vt_pt}
\end{figure}

The main results are shown in Table~\ref{tab:vt_pt}. For traditional virtual try-on, although our model is inferior in SSIM (please see the limitation discussion below for explanation), it achieves the best FID score. For pose transfer, our model is comparable with other SOTA models on all metrics. These results suggest that, although designed for MPVTON, our model is versatile and competitive for related tasks.  Some qualitative results are shown in Fig.~\ref{fig:fig_vt_pt}. 

\begin{table}[t]
    \centering
    \caption{Quantitative results of different traditional virtual try-on models on VTON benchmark and pose transfer models on MPV and DeepFashion testing split. \\ $\ast$ indicates that we trained (without distillation) PF-AFN on MPV and test it on VTON for reference}
    \setlength{\tabcolsep}{4pt}
    \begin{tabular}{c|c|c c c}
    \toprule
         \multirow{2}{*}{Tasks}&\multirow{2}{*}{Methods} & \multicolumn{3}{c}{Evaluation metrics}  \\
         \cline{3-5}
         & &FID $\downarrow$ & SSIM $\uparrow$& LPIPS $\downarrow$ \\
         \midrule
         
         \multirow{5}{*}{Traditional virtual try-on} &VTON \cite{han2018viton}&55.71&0.74&-\\
        &CP-VTON \cite{dong2019towards}&24.45&0.72&-\\
        &ACGPN
        \cite{yang2020towards}&16.64&0.84&-\\
        &Clothflow\cite{han2019clothflow}&14.43&0.84\\
        &$\text{PF-AFN}^{\ast}$ \cite{ge2021parser}&17.04&0.84&-\\
        &Zflow\cite{chopra2021zflow} &15.17&\textbf{0.88}\\
        &Ours&\textbf{13.51}&0.81&-\\
        \midrule
        
        \multirow{5}{*}{Pose transfer (MPV)} &ADGAN \cite{men2020controllable}&15.49&0.717&0.201\\
        &GFLA \cite{ren2020deep}&12.48&0.726&0.195\\
        &DiOr \cite{cui2021dressing}&13.33&\textbf{0.741}&\textbf{0.186}\\
        &Ours&\textbf{12.15}&0.738&0.187\\
        \midrule
        
        \multirow{5}{*}{Pose transfer (DeepFashion)} &ADGAN \cite{men2020controllable}&18.63&0.772&0.226\\
        &GFLA \cite{ren2020deep}&\textbf{13.18}&0.801&0.179\\
        &DiOr \cite{cui2021dressing}&13.59&\textbf{0.806}&\textbf{0.176}\\
        &Ours&14.36&0.801&0.184\\

    \bottomrule
    \end{tabular}
    \label{tab:vt_pt}
\end{table}

\begin{figure}[t!]
    \centering
    \includegraphics[width=0.5\textwidth]{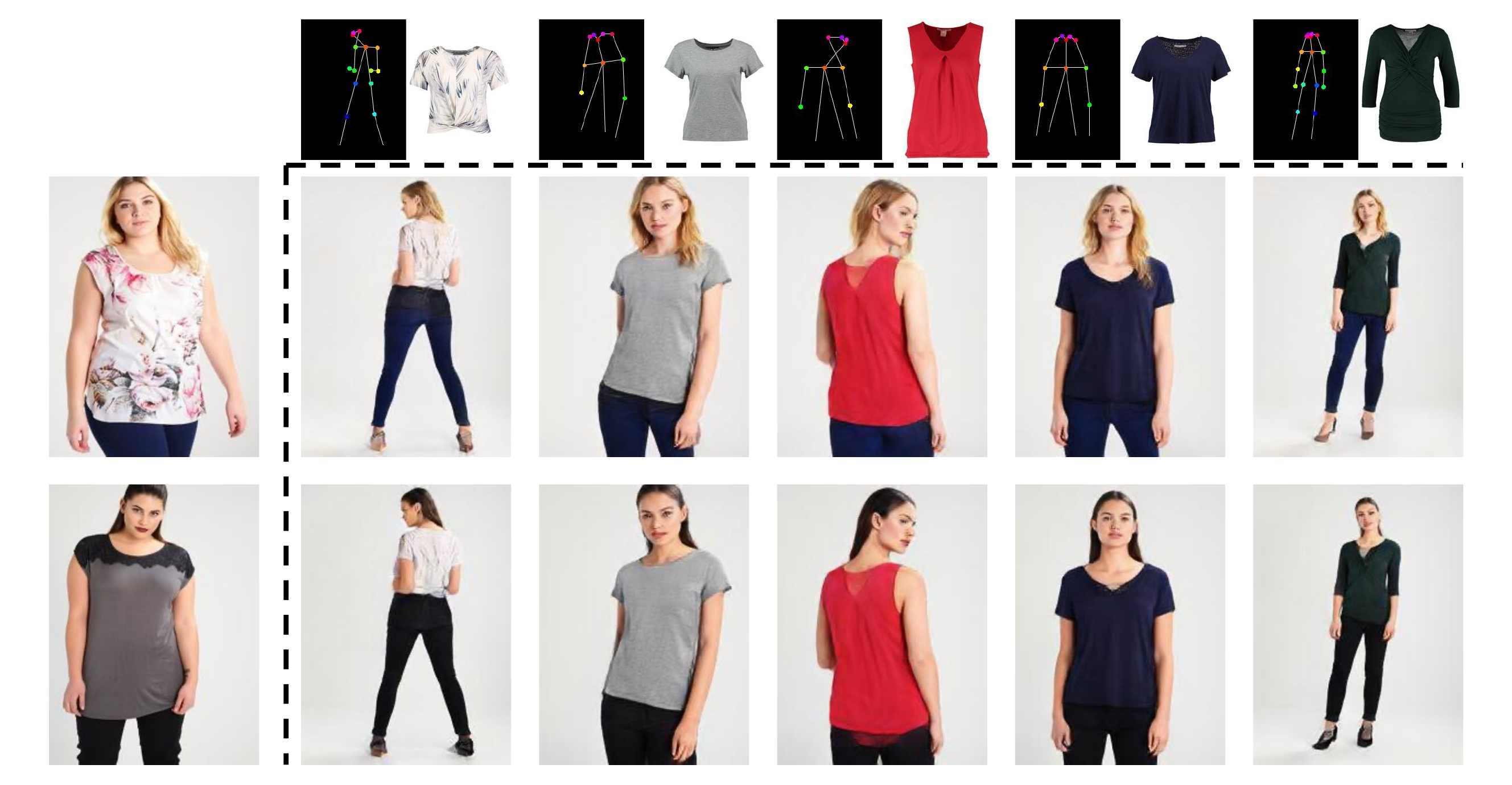}
    \caption{Illustrating the limitations of our model in body shape preservation.}
    \label{fig:fig_limi1}
\end{figure}

\begin{figure}[t!]
    \centering
    \includegraphics[width=0.5\textwidth]{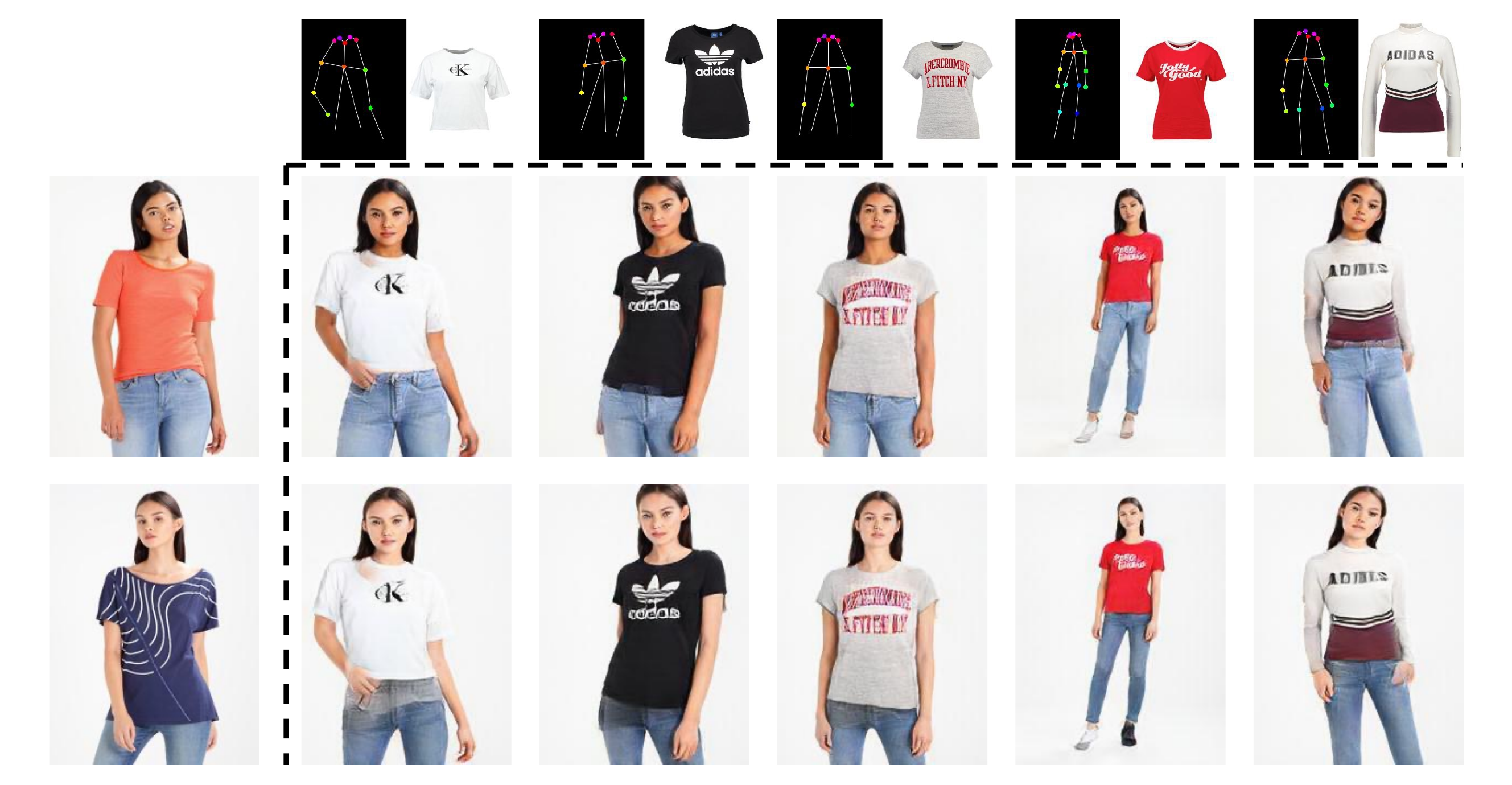}
    \caption{Illustrating the limitations of our model in garment texture preservation (please zoom in for better effect).}
    \label{fig:fig_limi2}
\end{figure}

\paragraph{\textbf{Limitations}} Although our model achieved state-of-the-art performance, it has scopes for improvements. In particular,  two main limitations are identified, namely body shape preservation and garment texture detail preservation. As illustrated in Fig.~\ref{fig:fig_limi1}, given a person image with a plus-size body shape, the person in the generated try-on image now has a slim body shape. This result suggests that our model failed to preserve body shape of the input person. This is likely caused by the dataset bias, as most person images in MPV are fashion models with slim body shapes. In the meantime, we found that our model sometimes struggles with garment detail preservation. As shown in Fig.~\ref{fig:fig_limi2}, the texture details of the garment can be distorted. This is especially true for the text printed on the garment. A plausible reason is  the bi-linear sampling based warping \cite{zhou2016view} which distorted the feature of each location. This also explains our lower SSIM score on the traditional virtual try-on task. For this task,  previous traditional try-on models directly use the raw images as input while our model uses the warped feature maps to guide the try-on image generation, so that it can have a chance to meet the main  MPVTON task's demand on large pose changes. 

\section{Conclusion}

In this paper, we have proposed a single-stage model for multi-pose virtual try-on. Our model uses target pose as condition to predict the flow fields for person and garment images and then warps their feature maps for target try-on image generation. Our model achieves state-of-the-art performance on the MPV benchmark and it is computationally more efficient compared to existing models. We conducted extensive experiments to show the superiority of our method and to validate our architecture design.

\bibliographystyle{IEEEtran}
\bibliography{ref.bib}

\vfill

\end{document}